\documentclass[10pt,twocolumn,letterpaper]{article}

\usepackage[pagenumbers]{cvpr} %

\usepackage{amsmath}
\usepackage{amssymb}
\usepackage{booktabs}
\usepackage{graphicx}       
\usepackage{subcaption}     
\usepackage{booktabs}
\usepackage{multirow}
\usepackage{bigstrut}
\usepackage{makecell}
\usepackage{array}
\usepackage{enumitem}
\usepackage{zhlipsum}
\newcommand{\tabincell}[2]{\begin{tabular}{@{}#1@{}}#2\end{tabular}}

\usepackage{xcolor,color,framed}
\usepackage{url, overpic, cases, contour}
\usepackage{multirow, makecell}

\definecolor{shadecolor}{rgb}{0.92,0.92,0.92}

\usepackage{adjustbox}
\makeatletter %
\@namedef{ver@everyshi.sty}{}
\makeatother
\usepackage{tikz}

\newlength \g

\usepackage[pagebackref,breaklinks,colorlinks]{hyperref}
\usepackage[symbol]{footmisc}

\usepackage[capitalize]{cleveref}
\crefname{section}{Sec.}{Secs.}
\Crefname{section}{Section}{Sections}
\Crefname{table}{Table}{Tables}
\crefname{table}{Tab.}{Tabs.}

\def\cvprPaperID{} %
\def\confName{CVPR}
\def\confYear{2022}

\begin{document}

\title{Compression-Realized Deep Structural Network for Video Quality Enhancement}
\author{Hanchi Sun$^{1}$ \qquad Xiaohong Liu$^{1}$ \qquad Xinyang Jiang$^{2}$  \qquad Yifei Shen$^{2}$\\ Dongsheng Li$^{2}$\qquad  \qquad Xiongkuo Min$^{1}$ \qquad Guangtao Zhai$^{1}$ \\
$^{1}$ Shanghai Jiao Tong University \quad
$^{2}$ Microsoft Research Asia\\
{\tt\small \{shc15522, xiaohongliu, minxiongkuo, zhaiguangtao\}@sjtu.edu.cn} \\ {\tt\small \{xinyangjiang, yifeishen, dongsli\}@microsoft.com}
}

\maketitle

\begin{abstract}
This paper focuses on the task of quality enhancement for compressed videos. 
Although deep network-based video restorers achieve impressive progress, most of the existing methods lack a structured design to optimally leverage the priors within compression codecs. Since the quality degradation of the video is primarily induced by the compression algorithm, a new paradigm is urgently needed for a more ``conscious'' process of quality enhancement. As a result, we propose the Compression-Realized Deep Structural Network (CRDS), introducing three inductive biases aligned with the three primary processes in the classic compression codec, merging the strengths of classical encoder architecture with deep network capabilities. 
Inspired by the residual extraction and domain transformation process in the codec, a pre-trained Latent Degradation Residual Auto-Encoder is proposed to transform video frames into a latent feature space, and the mutual neighborhood attention mechanism is integrated for precise motion estimation and residual extraction. Furthermore, drawing inspiration from the quantization noise distribution of the codec, CRDS proposes a novel Progressive Denoising framework with intermediate supervision that decomposes the quality enhancement into a series of simpler denoising sub-tasks.
Experimental results on datasets like LDV 2.0 and MFQE 2.0 indicate our approach surpasses state-of-the-art models.
\end{abstract}



\maketitle

\vspace{-2pt}

\section{Introduction}

In today's digital age, the need for transmitting high-quality videos over the Internet has surged. While video compression is pivotal for efficient video transmission given the bandwidth constraints of the Internet, it invariably introduces compression artifacts that can significantly diminish visual quality.

Over the past few years, a myriad of techniques based on deep learning have emerged to enhance the quality of compressed videos \cite{deng2020spatio, guan2019mfqe, huo2021recurrent, lu2018deep, wang2017novel, yang2018enhancing,chan2022basicvsr++}, which significantly outperforming classic methods.
However, existing deep statistical Video Quality Enhancement (VQE) models are largely black-box, overly dependent on the data distribution used for training, and lack a structured design to introduce inductive biases related to the video compression degradation model.

\begin{figure}[t!]
  \centering
  \subfloat[\label{fig:1-1}Codec Compression]{
    \includegraphics[width=8.5cm]{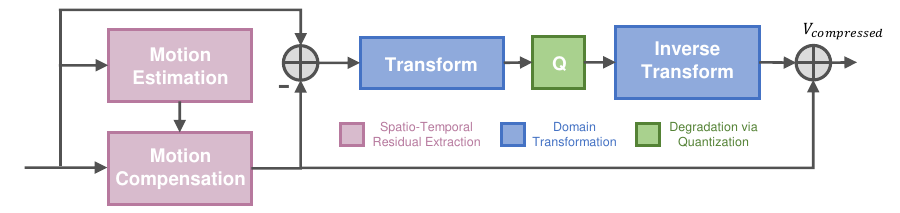}
  }
  
  \vspace{1em}

  \subfloat[\label{fig:1-2}CRDS]
  {
    \includegraphics[width=8.5cm]{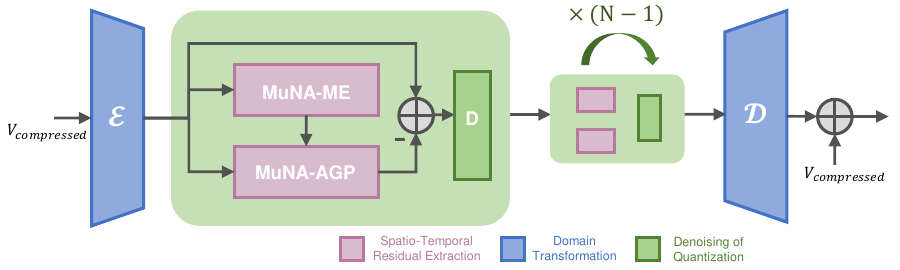}
  }
  
    \vspace{-6pt}
    
  \caption{The comparison between the Codec Compression framework and the CRDS framework. }
  \label{fig:fig1}
\vspace{-8pt}
\end{figure}

From a higher-level cognitive standpoint, inductive biases leverage priors to ``consciously'' assist in mobilizing various ``unconscious'' or black-box-like neural network modules to form a structured deep model, as proposed in \cite{goyal2022inductive}. This methodology underlies the development of models endowed with ``structural-mechanistic'' properties, adhering to a fundamental mechanistic understanding of reality, which is analogous to how a ``world simulator'' facilitates progress in Sora \cite{liu2024sora}. Existing deep VQE models lack a structured design aligned with codecs to better facilitate the introduction of compression-related inductive bias. 
While some studies briefly exploit the inherent temporal continuity of videos, none comprehensively explore all aspects of a compression degradation model. 
Therefore, there is a necessity for a new video quality enhancement paradigm that merges the strengths of classic codec architecture with deep network capabilities. 


In order to introduce sufficient compression-related inductive bias to form a deep structural VQE model, we start by first analyzing how noise is introduced in a degradation model resulting from compression. 
As shown in Figure \ref{fig:fig1}, in traditional video codecs, video compression typically undergoes three main processes: 1) 
\textbf{Spatio-Temporal Residual Extraction}: 
Based on spatial consistency for intra-frame prediction and leveraging temporal continuity for inter-frame prediction, block-wise motion vectors are estimated to predict the current block using the most similar areas in adjacent regions and frames, obtaining the residual between the original frame and the predicted frame; 2) \textbf{Domain Transformation}: Obtained frame residuals are converted from the data space to a latent space, such as the frequency domain or the wavelet domain, due to the latter's enhanced ability for better feature extraction and information separation; 3) \textbf{Degradation via Quantization}: 
The residuals obtained in the latent space are compressed by quantization and encoding.  

Consequently, the aforementioned prior knowledge from traditional codecs inspires us to introduce three inductive biases aligned with the three primary processes in the degradation model into the black-box-like deep statistical model, designing a more interpretable deep structured model, as shown in Figure \ref{fig:fig1}, from the perspectives of algorithms, network architecture, and training frameworks.

Inspired by Spatio-Temporal Residual Extraction, which involves aligning inter and intra frame blocks with motion vectors, we introduce a block-wise motion estimation algorithm called Mutual Neighborhood Attention (MuNA). By incorporating optical flow as guidance into the neighborhood attention mechanism, MuNA is capable of flexibly and more robustly revealing the associations between different blocks in both temporal and spatial dimensions, thereby more effectively estimating both intra and inter-frame motions at the macro-block level.

Following the Domain Transformation in codec degradation model, 
we propose 
a novel auto-encoder based pre-training framework called Latent Degradation Residual Auto-Encoder (LDR-AE) to represents the frame residual in a latent space. 
LDR-AE adopts a two-to-one architecture, wherein the encoder obtains the implicit representations of both high-quality (HQ) and low-quality (LQ) frames which are then simultaneously input into its decoder to predict the residual. 
Compared to conventional auto-encoders, LDR-AE reconstructs a residual rather than an entire image, and thus only requires a lightweight decoder to facilitate learning.
During the pre-training process, HQ frames with various levels of degradation are obtained by blending the original frame and compressed frame using different weight coefficients, which encourages LDR-AE to learn representations of compression across a spectrum of degradation levels.

In order to draw inspiration from the quantization process, we first analyze the distribution of the noises produced by residual quantization. 
We demonstrate that in the code space of video compression, quantizing residuals is equivalent to applying noises following a uniform distribution with the Quantization Parameter (QP) as its intensity, aptly referred to as quantization noise. 
Thus, the essence of video quality enhancement is the denoising process for quantization noise. 
More importantly, we demonstrate that a quantization process that applies higher-level noises can be effectively approximated as the accumulation of a series of lower-level noises. Thus, the entire video quality enhancement process can be decomposed into the sequential removal of multiple quantization noises of lesser intensity. 
As a result, we develop a progressive de-quantization network consisting of cascaded denoising blocks. Each block is trained to eliminate quantization noise of a particular lower intensity, utilizing the pre-trained decoder of LDR-AE for intermediate supervision. 
This approach, named Progressive Denoising via Intermediate Supervision (PDIS), compared to the single-step removal of complete quantization noise, results in more similar noisy inputs and partially denoised outputs, thereby reducing the blurring issues from using Euclidean distance for loss calculation.


To summarize, our CRDS introduces a new framework by incorporating video codec inductive bias into the network structure through novel designs in \textbf{algorithms (MuNA), training strategy (LDR-AE, PDIS), and network architecture (PDIS)}, which yields a more robust and generalizable deep structural VQE model compared to the more common deep statistical VQE models.

With extensive experiments, we demonstrate that the proposed method attains state-of-the-art results in video quality enhancement on the LDV 2.0 and MFQE 2.0 datasets, while also exhibiting superior out-of-distribution (OOD) generalization capabilities.
Furthermore, the comparison between our intermediate results and the metadata from the video bitstream confirms that the sequential sub-modules in our network possess interpretable functions that correspond to the video compression codec.

\section{Related Works}

\begin{figure*}[!htb]
  \centering
  \includegraphics[width=1\textwidth]{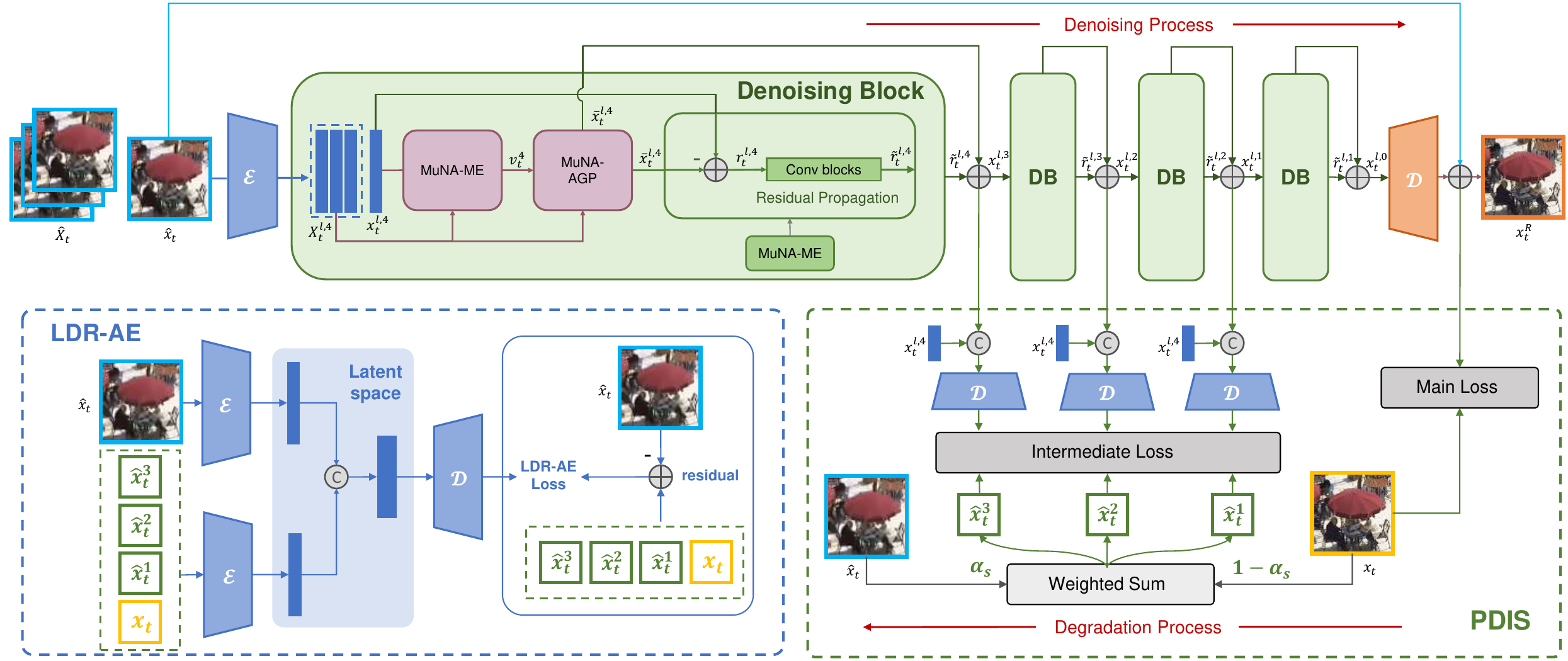} 
  \caption{An Overview of our proposed CRDS framework. The CRDS framework comprises three primary structures: MuNA, LDR-AE, and PDIS, which adequately introduced the inductive bias of video compression.}
  \label{fig:pipeline}
  \vspace{-4pt}
\end{figure*}

\subsection{Video Compression}

Recent decades have seen the introduction of advanced video compression standards such as H.264 \cite{wiegand2003overview} and H.265 \cite{sullivan2012overview}, which primarily employ predictive coding. This includes intra-frame and inter-frame predictions with essential motion estimation \cite{flierl2002rate}. The residuals, or differences between original and predicted frames, undergo Discrete Cosine Transform (DCT) \cite{ahmed1974discrete} and subsequent quantization \cite{ding1996rate}, filtering out less perceptible high-frequency information to achieve efficient compression.

\subsection{Video Quality Enhancement}
In the realm of video quality enhancement, different approaches have been explored. Wang et al. \cite{wang2017novel} and Yang et al. \cite{yang2018enhancing} concentrate on improving the quality of individual frames, while Guan et al. \cite{guan2019mfqe}, Huo et al. \cite{huo2021recurrent}, Deng et al. \cite{deng2020spatio}, and Lu et al. \cite{lu2018deep} focus on leveraging temporal correlations. For video super-resolution, Haris et al. \cite{haris2019recurrent} and Huang et al. \cite{huang2015bidirectional} address uncompressed video formats, whereas Chen et al. introduce a novel technique in 2020 tailored for compressed videos \cite{chen2020bitstream}, further refined in 2021 \cite{chen2021compressed}. By 2022, Chan et al. \cite{chan2022basicvsr++} developed BasicVSR++, an advanced network utilizing spatiotemporal data to enhance alignment and propagation in misaligned frames, marking a significant improvement in recurrent system capabilities.

Optimal utilization of temporal features is essential in video enhancement \cite{shi2021learning, liu2021exploit, liu2018robust, 10471308, li2023fastllve}, where alignment plays a critical role \cite{xue2019video, shi2021video, liu2020end, wu2024perception}. 
Xue et al. \cite{xue2019video} introduce SpyNet for video restoration, based on the techniques of optical flow estimation \cite{ranjan2017optical} and feature warping \cite{chan2021understanding, liu2017robust}.
Tian et al. \cite{tian2020tdan} and Wang et al. \cite{wang2019edvr} achieve better feature alignment utilizing deformable convolution \cite{dai2017deformable}. Building on this, Chan et al. \cite{chan2022basicvsr++} employ optical flow to guide deformable convolution.
Additionally, Liang et al. \cite{liang2022vrt} introduce a mutual attention mechanism for precise motion estimation, although the shifted window mechanism limits motion vectors' range \cite{liu2021swin}, and global self-attention has been found to be computationally inefficient \cite{han2022survey, arnab2021vivit}. Zhu et al. \cite{zhu2024cpga} leverage codec metadata for better VQE performance.

\subsection{Local Attention}

The Stand Alone Self Attention (SASA) model by Ramachandran et al. \cite{ramachandran2019stand} is an initial attempt to integrate sliding window self-attention into CNNs, offering improvements in accuracy but suffering from high latency. It influences subsequent models like Longformer \cite{beltagy2020longformer} and Vision Longformer (ViL) \cite{zhang2021multi}, which face challenges with scaling and limited receptive fields due to padding. Addressing these limitations, Liu et al. introduce the Swin Attention mechanisms \cite{liu2021swin}, which utilize non-sliding windows to partition feature maps and apply localized self-attention, effectively resolving previous issues. Building on existing attention mechanisms applied to various vision-related tasks \cite{shi2022video, wu2023accflow}, Hassani et al. develop the Neighborhood Attention (NA) \cite{hassani2023neighborhood}, the first efficient and scalable sliding window attention mechanism for visual tasks.

\section{Method}
\subsection{Overall Framework}

Given a sequence of low-quality video frames after compression $V^{LQ}=\{\widehat{x}_1, \widehat{x}_2, …, \widehat{x}_{t}, \widehat{x}_{t+1}, …\}$  encoded and decoded with a compression video codec from a sequence of the original video frames $V^{GT}=\{x_1, x_2, …, x_t, x_{t+1}, …\}$, where $x_t$ represents the $t$-th frame, this paper focuses on restore and enhance the quality of $V^{LQ}$ to obtain a high-quality frame sequence close to $V^{GT}$.  

The compression process of a video codec process follows three key steps, as is elaborated below. 

\begin{itemize}[leftmargin=0pt, itemindent=2ex, topsep=1pt, itemsep=0pt, partopsep=0pt]

\item
\textbf{Spatiotemporal Residual Extraction. }
Firstly, Block-wise motion estimation is performed within the current frame $x_t$ and adjacent reconstructed frames $\widehat X_t$, to obtain the block-wise motion vector $v_t$. 
Secondly, Intra-frame prediction of $x_t$ and Inter-frame prediction between $x_t$ and $\widehat X_t$ with the help of $v_t$ are simultaneously implemented, to obtain the optimal predicted frame $\bar{x}_t$ and extract the residual between $x_t$ and $\bar{x}_t$:
\begin{equation}
{{v}_t} = ME({{x}_t},{\widehat X_t}),
\end{equation}
\begin{equation}
({\overline x _t},{r_t}) = P({x_t},{\widehat X_t},{v_t}).
\end{equation}

\item
\textbf{Domain Transformation. }
The residual $r_t$ undergoes a certain transformation from the data space into the code space:
\begin{equation}
r_t^c = T({r_t}),
\end{equation}
where $r_t^c$ denotes the residual in the code space.

\item
\textbf{Degradation. }
$r_t^c$ is compressed in a lossy manner through quantization, resulting in $\widehat{r}_t^c$:
\begin{equation}
{\widehat r_t}^c = Q(r_t^c).
\label{eq:quantization}
\end{equation}
Then the quantized residual $\widehat{r}_t$ in the data space is obtained through inverse transformation and added back to $\bar{x}_t$, resulting in the compressed reconstructed frame $\widehat{x}_{t}$ that forms $V^{LQ}$:
\begin{equation}
{\widehat x_t} = {\overline x _t} + {T^{ - 1}}({\widehat r_t}^c) = {\overline x _t} + {\widehat r_t}.
\label{eq:reconstruct}
\end{equation}

\end{itemize}

Based on the analysis of the video compression process, we propose a Compression-Realized Deep Structural network (CRDS), wherein reasonable inductive biases are introduced through the three novel designs, namely MuNA, LDR-AE, and PDIS. 

As shown in Figure \ref{fig:pipeline}, different from conventional codec algorithms, CRDS conducts the video quality enhancement in a latent feature space from the beginning, where more semantic information will be embedded. The low-quality frames $\widehat{x}_{t}$ and $\widehat{X}_{t}$ are transformed into the latent representations $x^l_{t}$ and $X^l_{t}$ by the feature encoder of LDR-AE, which is pre-trained on compressed videos with self-reconstruction loss to better facilitate the introduction of compression-related inductive bias. The detailed implementation and training of LDR-AE will be introduced in Section \ref{sec:ldr-ae}. 
In the latent space, CRDS conducts a progressive denoising process called PDIS with a series of denoising blocks trained with a novel progressive intermediate supervision, each of which removes a specific part of the lower-level quantization noises from the LQ videos, enhancing $x^{l,s}_{t}$ from the preceding block to $x^{l,s-1}_{t}$. The detailed implementation of PDIS will be introduced in Section \ref{sec:pdis}.   
Specifically, each denoising block in PDIS removes the quantization noise by first obtaining the motion vectors $v_t^l$ for residual extraction with Mutual Neighborhood Attention (MuNA). The obtained residuals $r_t^l$ are then denoised and enhanced by propagating temporal features across multiple frames in the Residual Propagation Network, producing $\widetilde r_t^l$. Details of the denoising block and MuNA will be introduced in Section \ref{sec:muna}. 

\subsection{Latent Degradation Residual Auto-Encoder} \label{sec:ldr-ae}
\textbf{Two-to-one encoder-decoder architecture. }
In the LDR-AE model, the encoder comprises a series of consecutive residual blocks to convert video frames from the data space to the latent space, while simultaneously extracting deep spatial features from individual images. The low-quality image, $x_{lq}$, and the high-quality image, $x_{hq}$, are processed through the encoder to generate their latent representations in the high-dimensional space, $x_{lq}^l$ and $x_{hq}^l$, respectively. The decoder then combines $x_{lq}^l$ and $x_{hq}^l$ and reconstructs the degradation residual $r_d$ between them in the data space. 
\begin{equation}
({x_{lq}^l},{x_{hq}^l}) = \mathcal{E}({x_{lq}},{x_{hq}})
\end{equation}
\begin{equation}
{r_d} = \mathcal{D}({x_{lq}^l},{x_{hq}^l})
\end{equation}
If in different pairs of low-quality and high-quality images, $x_{lq}$ is consistently derived from $x_{hq}$ through the same type of degradation, then it is reasonable to believe that LDR-AE is in pursuit of a high-dimensional latent space that not only can comprehensively represent the original image implicitly but also extract deep features oriented towards a specific degradation. In other words, LDR-AE has learned something fundamental and generalizable about the nature of the degradation.

\vspace{3pt}
\noindent
\textbf{Multi-level degradation residuals. }
When constructing the training set, we mix a compressed frame $\widehat{x_t}$ and the pre-compression original frame $x_t$ at various ratios to create partially degraded images as $\widehat x_t^s$, to pair with the fully compressed image $\widehat{x_t}$ as inputs for LDR-AE, which can be denoted as 
\begin{equation}
\widehat x_t^s = \alpha _s \cdot {\widehat x_t} + (1 - \alpha _s) \cdot {x_t}.
\label{eq:hybrid}
\end{equation}
\begin{equation}
({x_{lq}},{x_{hq}}) := (\widehat{x}_t, \widehat{x}_t^s). 
\end{equation}
``:='' means that $(x_{lq}, x_{hq})$ is assigned the value of $(\widehat{x}_t, \widehat{x}_t^s)$. To ensure training stability, we selected several fixed degradation levels: $\alpha _s = 0, 0.25, 0.5, 0.75$ for $s = 0, 1, 2, 3$. 

Consequently, the target reconstructed by the decoder is a degradation residual of multi-levels. In practice, the degree of a certain type of degradation is often determined by different noise level coefficients, such as the degree of video compression being determined by the quantization parameter (QP). Through this method of data augmentation via mixing, we can aid LDR-AE in better learning the generative mechanism of a certain form of degradation.

\vspace{3pt}
\noindent
\textbf{Multi-level training objective. }
We desire LDR-AE to possess progressive learning capacities for degradation of higher levels, thereby enhancing its ability to learn the full spectrum of degradation introduced by complete video compression. Accordingly, we implement reconstruction losses with variable weight coefficients for inputs associated with diverse degradation levels:
\begin{equation}
\mathcal{L} = \frac{{\left\| {\theta ({\widehat{x_t}},{\widehat{x_t}}^s)) + {\widehat{x_t}} - {\widehat{x_t}}^s} \right\| _2^2}}{{\alpha _s + \varepsilon }}
\end{equation}
where $\theta$ denotes the LDR-AE network and $\varepsilon$ is a predefined bias constant to prevent an overly small denominator in the loss function. During the pre-training process of LDR-AE, we set $\varepsilon=0.25$.

\subsection{Progressive Denoising via Intermediate Supervision}
\label{sec:pdis}

\textbf{Degradation process. }
As demonstrated in \cite{balle2016end}, the degradation caused by quantization can be equivalently regarded as additive uniform noise on the residual in the code space. Therefore, Equation \ref{eq:quantization} can be rewritten as:
\begin{equation}
{\widehat r_t}^c = r_t^c + \eta,
\label{eq:q}
\end{equation}
where $\eta$ represents the quantization noise that is derived from the residual calculation.
                               
Given a video frame $x_t^c$ in the code space, consider the degradation operator $D$ at severity level $s$, where $\alpha_s$ signifies the intensity of quantization, expressed as:
\begin{equation}
D(x_t^c,s) = x_t^c + {\alpha _s} \cdot \eta,
\label{eq:d}
\end{equation}

When we denote $s \in [0, N]$ as the discrete time step, the noise increment between two adjacent noise levels can be expressed as:
\begin{equation}
d(x_t^c,s) = D(x_t^c,s) - D(x_t^c,s-1) = {\beta _s} \cdot \eta,
\label{eq:noise}
\end{equation}
where $\beta _s=\alpha _s-\alpha _{s-1}$. The noise ${\beta _s} \eta$ can be considered as a lower-level quantization noise. Consequently, quantization noise $\eta$ can be decomposed into a sequential accumulation of ${\beta _s} \eta$ from $s=0$ to $s=N$, which constitutes the degradation process of video quantization.

We use $\alpha _s$ as the weighting coefficient to blend the original frame $x_t$ and the compressed frame $\hat{x}_t$ into $\widehat x_t^s$, which follows the same generation method as in Equation \ref{eq:hybrid}. $\widehat x_t^s$ serves as a progressive hybrid target \cite{ren2023deepmim} for intermediate supervision.

Given that the domain transformation in the codec is linear, this process of generating a progressive hybrid target in the data space is equivalent to the degradation process in the code space as described by Equation \ref{eq:d}.

\vspace{3pt}
\noindent
\textbf{Denoising process. }
CRDS adopts multiple denoising blocks for progressive video enhancement due to two main reasons. Firstly, the lack of access to the original high-quality frame hinders accurate residual approximation, necessitating progressive refinement of the residual $r_t^{l,s}$ through advanced neural network architectures. Secondly, by performing stage-wise partial denoising of lower-level noise instead of single-step removal of higher-level quantization noise, the model ensures that the process of residual estimation and enhancement progresses positively, leading to more precise and constrained video quality enhancement.

The denoising block consists of two main parts: residual estimation and residual enhancement. Residual estimation is implemented by the MuNA network. For residual enhancement, we propose a Residual Propagation Network to enhance the frame residual obtained by MuNA, with special residual connections and a recurrent structure. The residual $r_t^{l,s}$ is enhanced to $\widetilde r_t^{l,s}$ by propagating temporal features across multiple frames. The residual enhancement in the Residual Propagation Network can be formulated as:
\begin{equation}
    \widetilde r_t^{l,s} - r_t^{l,s} = - \eta ^{l,s} \sim  - {\beta _s} \cdot \eta.
\label{eq:denoise}
\end{equation}
$\eta ^{l,s}$ refers to the noise removed by the Residual Propagation Network $\Phi _{RP}^s$, where $s$ is the stage number of the denoising block. As shown in Equation \ref{eq:denoise}, we aim for $\eta ^{l,s}$ in the latent space to correspond closely to ${\beta _s} \cdot \eta$ in the code space, thereby realizing the inverse of the degradation process as described in Equation \ref{eq:noise}.

At the end of the Residual Propagation Network, the enhanced latent residual $\widetilde r_t^{l,s}$ is added to the predicted frame $\bar x_t^{l,s}$ and fed into subsequent denoising blocks as the partially denoised $x_t^{l,s-1}$ for further enhancement, as shown in Figure \ref{fig:pipeline}.

\begin{figure}
  \centering
  \includegraphics[width=0.47\textwidth]{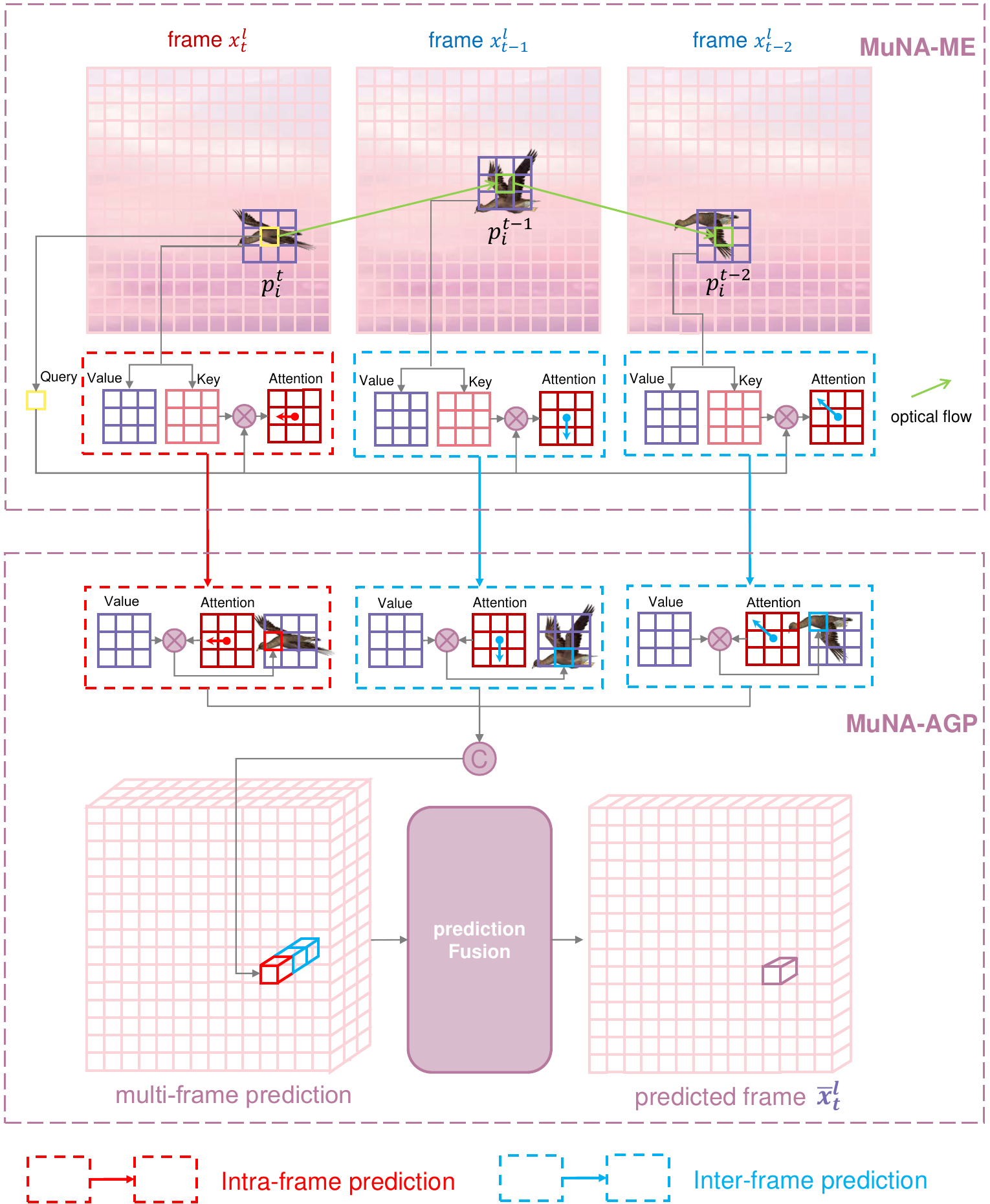} 
  \caption{A visual demonstration of Mutual Neighborhood Attention. MuNA-ME estimates motion vectors by performing mutual attention on two frames. MuNA-AGP carries out feature warping using attention weights to produce the predicted frame.}
  \label{fig:fig3}
  \setlength{\belowcaptionskip}{-30pt} 
\vspace{-20pt}
\end{figure}

The entire denoising block can be formulated as:
\vspace{-2pt}
\begin{equation}
\left\{ {\begin{array}{*{20}{c}}
{\Phi _{MuNA}^s(x_t^{l,s}, X_t^{l,s}) = r_t^{l,s}}\\
{\Phi _{RP}^s(r_t^{l,s}) = \widetilde r_t^{l,s} + \bar x_t^{l,s}} = x_t^{l,s-1}
\end{array}} \right.
\label{eq:denoising block}
\vspace{-2pt}
\end{equation}
where $\Phi _{MuNA}^s$ denotes the MuNA network in the denoising block of stage $s$. Since denoising is the inverse process of degradation, the stage $s$ to which the denoising blocks belong decreases from $N$ to 0, and the final enhanced frame is reconstructed by $x_t^{l,0}$. We set $N=4$ in our network, as shown in Figure \ref{fig:pipeline}.

\vspace{3pt}
\noindent
\textbf{Progressive training objective. }
We connect the output of the denoising block to the lightweight decoder of the pre-trained model LDR-AE to reconstruct the enhanced frame in the data space and train it against the progressive hybrid target as a reference. 

We train the CRDS network in three stages: 1) \textbf{In the first stage,} the encoder provided by the pre-trained LDR-AE is frozen, and the loss function is a weighted sum of the intermediate loss $L_I$ and the main loss $L_M$, where $L_I$ is an L2 loss and $L_M$ is a Charbonnier loss; 2) \textbf{In the second stage,} the encoder module is unfrozen; 3) \textbf{In the third stage,} training is performed only using the main loss $L_M$. 
{
\begin{small}
\vspace{-4pt}
\begin{gather}
  \mathcal{L_I} = \sum\limits_{s = 1}^{N - 1} {\left\| {\mathcal{D}(x_t^{l,N},x_t^{l,s}) - \widehat x_t^s} \right\| _2^2} . \\
  \addlinespace[-4pt] 
  \mathcal{L_M} = Charbonnier\left( {\mathcal{D}(x_t^{l,N},x_t^{l,0}),  x_t} \right).
\end{gather}
\vspace{-13pt}
\end{small}
}

Freezing the encoder in the first stage helps leverage the latent space learned by LDR-AE, tailored for representing quantization degradation, to constrain the early learning stage. The Intermediate loss function $L_I$ aims to constrain the specific quantization noise that each stage's denoising block needs to remove, effectively limiting the enhancement path. Subsequently, unfreezing the encoder and only applying the main loss function $L_M$ allows for the fine-tuning of the entire end-to-end network to optimize VQE results.

\subsection{Mutual Neighborhood Attention}
\label{sec:muna}

CRDS proposes a mutual neighborhood attention module for motion estimation (MuNA-ME). 
As shown in Figure \ref{fig:fig3}, for each sub-block in a frame, MuNA-ME aims to find a motion vector pointing to the most relevant sub-blocks in a group of reference frames $\{x_{t+m}^l | m=-m_w, -m_w+1, ..., m_w\}$.  
The reference frame could be the adjacent frames of the current frame (i.e. inter-frame motion vector, where $m\neq0$), or it could also be the current frame the sub-block belongs to (i.e. intra-frame motion vector, where $m=0$). 

Given the $i$-th sub-block $p_i^t$ in $x^l_t$, to obtain its motion vector towards the $m$-th reference frame $x^l_{t+m}$, we use optical flow to locate a reference window $w_i^{t+m}$ containing $k$ candidate sub-blocks. 
To more precisely locate the target sub-blocks most relevant to $p_i^t$ within the candidate window, MuNA-ME adopts an attention-based method to obtain the similarity between $p_i^t$ and sub-blocks in the reference window $w_i^{t+m}$. 

Firstly, following the common practice of the self-attention method, we treat the target sub-block $p_i^t$ as query, and obtain its attention on the candidate sub-blocks in the candidate window. 
Specifically, the query features $Q^t$  is obtained by applying linear projection on the current frame  $H^t$, and the key features is obtained from the reference frames  $H^{t+m}$, which is formulated as follows: 
\begin{equation}
Q^t=H^t P^Q, \quad K^{t+m}=H^{t+m} P^K,
\end{equation}

\begin{equation}
\mathbf{A}_i^{(t,t+m)}=\left[\begin{array}{c}
Q_i^t (K_{\rho_1(i)}^{t+m})^T+B{\left(i, \rho_1^{t+m} (i)\right)} \\
Q_i^t (K_{\rho_w(i)}^{t+m})^T+B{\left(i, \rho_2^{t+m} (i)\right)} \\
\vdots \\
Q_i^t (K_{\rho_k(i)}^{t+m})^T+B{\left(i, \rho_k^{t+m} (i)\right)}
\end{array}\right]
\end{equation}
where $\rho_j^{t+m} (i)$ denotes $p_i^{t+m}$’s $j$-th nearest neighbor. 

\begin{table*}[htbp]
  \setlength\tabcolsep{3pt} 
  \centering
  \caption{Performance comparison between CRDS and existing video restoration or enhancement methods on LDV 2.0 \cite{yang2022ntire} and MFQE 2.0 \cite{guan2019mfqe} in terms of the number of parameters, runtime, $\Delta$PSNR and $\Delta$SSIM. [Key: \textcolor{red}{Best}; 
  \textcolor{blue}{Second Best}].}
  \vspace{-8pt}
  \scriptsize 
    \begin{tabular}{l|c|c||c|p{0.57cm}<{\centering}p{0.57cm}<{\centering}p{0.57cm}<{\centering}p{0.57cm}<{\centering}p{0.57cm}<{\centering}p{0.57cm}<{\centering}p{0.57cm}<{\centering}p{0.57cm}<{\centering}p{0.57cm}<{\centering}p{0.57cm}<{\centering}p{0.57cm}<{\centering}p{0.57cm}<{\centering}p{0.57cm}<{\centering}p{0.57cm}<{\centering}p{0.57cm}<{\centering}|p{0.57cm}<{\centering}||p{0.57cm}<{\centering}}
    \toprule
    \multicolumn{1}{l|}{Method} & \multicolumn{1}{l|}{\tabincell{c}{Params\\(M)}} & \multicolumn{1}{c||}{\tabincell{c}{Runtime\\(ms)}} & \multicolumn{1}{c|}{Metrics} & \multicolumn{1}{c}{\#1} & \multicolumn{1}{c}{\#2} & \multicolumn{1}{c}{\#3} & \multicolumn{1}{c}{\#4} & \multicolumn{1}{c}{\#5} & \multicolumn{1}{c}{\#6} & \multicolumn{1}{c}{\#7} & \multicolumn{1}{c}{\#8} & \multicolumn{1}{c}{\#9} & \multicolumn{1}{c}{\#10} & \multicolumn{1}{c}{\#11} & \multicolumn{1}{c}{\#12} & \multicolumn{1}{c}{\#13} & \multicolumn{1}{c}{\#14} & \multicolumn{1}{c|}{\#15} & \multicolumn{1}{c||}{Ave.} & \multicolumn{1}{c}{\tabincell{c}{MFQE\\2.0}} \bigstrut\\
    \midrule
    \multicolumn{1}{l|}{\multirow{2}*{EDVR \cite{wang2019edvr}}} & \multirow{2}*{20.6}  & \multirow{2}*{378}   & $\Delta$PSNR  &  1.08 & 2.03 & 1.95 & 2.22 & 1.45 & 1.36 & 1.41 & 1.86 & 1.61 & 1.40 & 0.52 & 0.80 & 1.75 & 1.48 & 1.87  &   1.52    &  0.77 \bigstrut[t]\\
          &       &       & $\Delta$SSIM  & 0.013 & 0.037 & 0.028 & 0.031 & 0.070 & 0.023 & 0.035 & 0.035 & 0.036 & 0.021 & 0.039 & 0.014 & 0.025 & 0.015 & 0.027 & 0.030 & 0.018
 \bigstrut[b]\\
    \hline
    \multicolumn{1}{l|}{\multirow{2}*{VSRT \cite{deng2020spatio}}} & \multirow{2}*{32.6}  & \multirow{2}*{328}   & $\Delta$PSNR  &  0.91 & 2.03 & 2.00 & 2.27 & 1.46 & 1.39 & 1.24 & 2.02 & 1.53 & 1.25 & 0.42 & 0.84 & 1.77 & 1.69 & 1.84 &   1.51    &  0.87 \bigstrut[t]\\
          &       &       & $\Delta$SSIM  &   0.008 & 0.034 & 0.030 & 0.031 & 0.070 & 0.024 & 0.033 & 0.037 & 0.033 & 0.019 & 0.041 & 0.016 & 0.028 & 0.016 & 0.034  &   0.030    &  0.020 \bigstrut[b]\\
    \hline
    \multicolumn{1}{l|}{\multirow{2}*{VRT \cite{liang2022vrt}}} & \multirow{2}*{30.7}  & \multirow{2}*{236}   & $\Delta$PSNR  &  1.12 & 2.05 & 2.09 & 2.37 & 1.52 & \textcolor{blue}{1.51} & \textcolor{blue}{1.51} & 2.14 & 1.70 & 1.65 & 0.69 & 0.90 & 1.67 & 1.68 & \textcolor{blue}{2.15}  &   1.65    &  1.06 \bigstrut[t]\\
          &       &       & $\Delta$SSIM  & 0.009 & 0.035 & 0.031 & 0.032 & 0.071 & 0.025 & 0.034 & 0.038 & 0.034 & 0.020 & 0.042 & 0.017 & 0.029 & 0.017 & 0.035 & 0.031 &  0.024
  \bigstrut[b]\\
    \hline
    \multicolumn{1}{l|}{\multirow{2}*{RVRT \cite{liang2022recurrent}}} & \multirow{2}*{10.8}  & \multirow{2}*{123}   & $\Delta$PSNR  &1.22 & 2.27 & 2.06 & 2.41 & 1.65 & 1.37 & 1.46 & 2.04 & 1.79 & 1.49 & \textcolor{blue}{0.78} & \textcolor{blue}{1.04} & \textcolor{blue}{1.81} & 1.61 & 2.07   &   1.67    & 1.08 \bigstrut[t]\\
          &       &       & $\Delta$SSIM  & 0.015 & 0.042 & \textcolor{blue}{0.044} & 0.045 & 0.076 & 0.033 & 0.036 & 0.042 & 0.038 & 0.022 & \textcolor{blue}{0.053} & 0.020 & 0.032 & 0.019 & 0.039 & 0.037 & 0.027
  \bigstrut[b]\\
    \hline
    \multicolumn{1}{l|}{\multirow{2}*{BasicVSR \cite{chan2021basicvsr}}} & \multirow{2}*{6.3}   & \multirow{2}*{63}    & $\Delta$PSNR  & 1.23 & 2.08 & 1.86 & 2.19 & 1.63 & 1.37 & 1.37 & 2.11 & 1.70 & 1.49 & 0.57 & 0.76 & 1.71 & 1.65 & 1.95 &   1.58    & 1.05 \bigstrut[t]\\
          &       &       & $\Delta$SSIM  & 0.015 & 0.041 & 0.038 & 0.038 & 0.074 & 0.031 & 0.035 & 0.037 & 0.034 & 0.027 & 0.045 & 0.029 & 0.036 & 0.019 & 0.035 & 0.036 & 0.025
  \bigstrut[b]\\
    \hline
    \multicolumn{1}{l|}{\multirow{2}*{IconVSR \cite{chan2021basicvsr}}} & \multirow{2}*{8.7}   & \multirow{2}*{70}    & $\Delta$PSNR  &1.25 & 2.00 & 2.04 & 2.30 & 1.66 & 1.30 & \textcolor{blue}{1.51} & 2.24 & \textcolor{blue}{1.83} & 1.81 & 0.54 & 1.00 & 1.66 & \textcolor{blue}{1.77} & 1.98 &   1.66    &  1.08 \bigstrut[t]\\
          &       &       & $\Delta$SSIM  & 0.017 & 0.041 & 0.043 & 0.041 & \textcolor{blue}{0.079} & 0.033 & \textcolor{blue}{0.040} & \textcolor{blue}{0.044} & 0.041 & 0.022 & 0.050 & \textcolor{blue}{0.033} & \textcolor{blue}{0.040} & \textcolor{blue}{0.031} & 0.038 & 0.037 & \textcolor{blue}{0.033}
  \bigstrut[b]\\
    \hline
    \multicolumn{1}{l|}{\multirow{2}*{BasicVSR++\cite{chan2022basicvsr++}}} & \multirow{2}*{7.3}   & \multirow{2}*{77}    & $\Delta$PSNR  &  \textcolor{blue}{1.35} & \textcolor{blue}{2.31} & \textcolor{blue}{2.11} & \textcolor{blue}{2.61} & \textcolor{blue}{1.69} & 1.36 & 1.49 & \textcolor{blue}{2.28} & 1.73 & \textcolor{blue}{1.82} & 0.61 & 1.02 & 1.79 & 1.71 & 2.08 &  \textcolor{blue}{1.73}     &  \textcolor{blue}{1.10} \bigstrut[t]\\
          &     &     & $\Delta$SSIM  & \textcolor{blue}{0.019} & \textcolor{blue}{0.047} & 0.040 & \textcolor{blue}{0.046} & 0.077 & \textcolor{blue}{0.034} & 0.038 & 0.039 & \textcolor{blue}{0.042} & \textcolor{blue}{0.031} & 0.050 & 0.024 & 0.037 & 0.025 & \textcolor{blue}{0.041} & \textcolor{blue}{0.039} &  0.032
  \bigstrut[b]\\
    \hline
    \multicolumn{1}{l|}{\multirow{2}*{\textbf{CRDS}}} & \multirow{2}*{8.1}   & \multirow{2}*{85}    & $\Delta$PSNR  &  \textcolor{red}{\textbf{1.59}} & \textcolor{red}{\textbf{2.51}} & \textcolor{red}{\textbf{2.56}} & \textcolor{red}{\textbf{2.75}} & \textcolor{red}{\textbf{2.11}} & \textcolor{red}{\textbf{1.72}} & \textcolor{red}{\textbf{1.87}} & \textcolor{red}{\textbf{2.53}} & \textcolor{red}{\textbf{2.07}} & \textcolor{red}{\textbf{2.02}} & \textcolor{red}{\textbf{0.96}} & \textcolor{red}{\textbf{1.56}} & \textcolor{red}{\textbf{2.14}} & \textcolor{red}{\textbf{2.01}} & \textcolor{red}{\textbf{2.51}}  &  \textcolor{red}{\textbf{2.06}}   &  \textcolor{red} {\textbf{1.33}} \bigstrut[t]\\
          &      &      & $\Delta$SSIM  & \textcolor{red}{\textbf{0.030}} & \textcolor{red}{\textbf{0.054}} & \textcolor{red}{\textbf{0.053}} & \textcolor{red}{\textbf{0.053}} & \textcolor{red}{\textbf{0.095}} & \textcolor{red}{\textbf{0.046}} & \textcolor{red}{\textbf{0.048}} & \textcolor{red}{\textbf{0.051}} & \textcolor{red}{\textbf{0.050}} & \textcolor{red}{\textbf{0.041}} & \textcolor{red}{\textbf{0.062}} & \textcolor{red}{\textbf{0.039}} & \textcolor{red}{\textbf{0.050}} & \textcolor{red}{\textbf{0.034}} & \textcolor{red}{\textbf{0.052}} & \textcolor{red} {\textbf{0.051}} & \textcolor{red} {\textbf{0.042}} 
  \bigstrut[b]\\
    \bottomrule
    \end{tabular}%
  \normalsize 
  \label{tab:table1}%
\end{table*}%

Eventually, the motion vector between the current frame $x^l_t$ and the reference frame $x^l_{t+m}$ estimated by MuNA-ME is the sum of optical flow offsets and the vector representations of attention weights:
\begin{equation}
 MV^{(t,t+m)} = OF^{(t,t+m)} + Vector\left(A^{(t,t+m)}\right)
\end{equation}


We further introduce the mutual neighborhood attention for attention-guided frame prediction (MuNA-AGP), which exploits the motion vectors from MuNA-ME to conduct frame prediction.
We first obtain the value representation of the reference frame $V^{t+m}$  by a linear projection on $H^{t+m}$: 
\begin{equation}
 V^{t+m}=H^{t+m} P^v,
\end{equation}

To predict a sub-block $\bar{p}_i^t$ in a video frame, we leverage the obtained motion vector $\mathbf{A}_i^{(t,t+m)}$ to warp the value projections of $k$ nearest neighboring sub-blocks $p_i^{t+m}$ within the reference window, in a manner of weighted average:
 

\begin{equation}
\mathrm{\bar{H}}^{(t,t+m)}(i)=\operatorname{softmax}\left(\frac{\mathbf{A}_i^{(t,t+m)}}{\sqrt{d}}\right) \mathbf{V}_i^{t+m},
\end{equation}
where, 
\begin{equation}
\mathbf{V}_i^{t+m}=\left[\begin{array}{llll}
(V_{\rho_1^{t+m} (i)}^{t+m})^T & (V_{\rho_2^{t+m} (i)}^{t+m})^T & \ldots & (V_{\rho_k^{t+m} (i)}^{t+m})^T
\end{array}\right]^T, 
\label{eq:frame_prediction}
\end{equation}
$\sqrt{d}$ is the scaling parameter. This operation is conducted on every patch in the feature map.

We propose a prediction fusion network containing a group of feature fusion blocks, which merges the intra and inter-frame prediction for the final predicted frame $\bar{x}_t^l$. Ultimately, we extract the residual of $x_t^l$ through MuNA:
\begin{equation}
    r_t^l = x_t^l - \bar{x}_t^l.
\end{equation}

\section{Experiments}
\subsection{Experimental Setup}

\begin{figure*}
  \centering
  \includegraphics[width=0.95\textwidth]{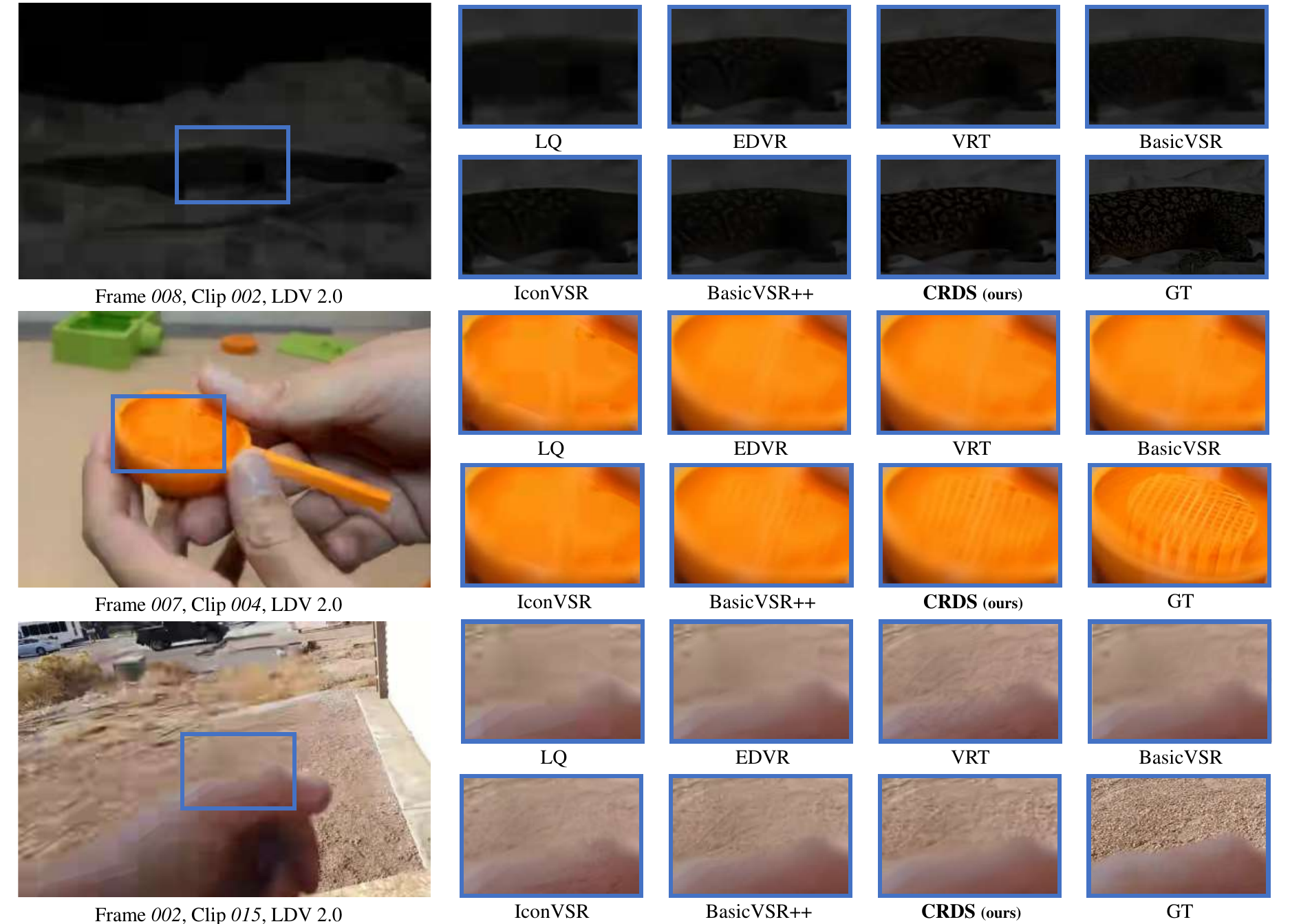} 
    \vspace{-6pt}
  \caption{Qualitative comparison between CRDS and existing SOTA baselines on \textbf{LDV 2.0 \cite{yang2022ntire}}. }
  \label{fig:fig4}
  \vspace{-4pt}
\end{figure*}

\begin{table*}[htbp]
  \centering
  \caption{{Performance comparison between CRDS and existing video enhancement methods when the model is trained on MFQE 2.0 \cite{guan2019mfqe} and tested on LDV 2.0 \cite{yang2022ntire} in terms of $\Delta$PSNR and $\Delta$SSIM. [Key: \textcolor{red}{Best}; \textcolor{blue}{Second Best}].}}
    \vspace{-7pt}
  \footnotesize 
    \begin{tabular}{p{1.59cm}<{\centering}|p{1.59cm}<{\centering}|p{1.59cm}<{\centering}|p{1.59cm}<{\centering}|p{1.59cm}<{\centering}|p{1.59cm}<{\centering}|p{1.59cm}<{\centering}|p{1.59cm}<{\centering}||p{1.59cm}<{\centering}}
    \hline
    Method &  \multicolumn{1}{c|}{EDVR \cite{wang2019edvr}} & \multicolumn{1}{c|}{VSRT \cite{deng2020spatio}} & \multicolumn{1}{c|}{VRT \cite{liang2022vrt}} & \multicolumn{1}{c|}{RVRT \cite{liang2022recurrent}} & \multicolumn{1}{c|}{BasicVSR \cite{chan2021basicvsr}} & \multicolumn{1}{c|}{IconVSR \cite{chan2021basicvsr}} & \multicolumn{1}{c||}{BasicVSR++\cite{chan2022basicvsr++}} & \multicolumn{1}{c}{\textbf{CRDS}} \bigstrut\\
    \hline
    $\Delta$PSNR (dB)  &   0.53    &   0.59    &     0.71  &    0.67   &   0.86    &   0.87    &   \textcolor{blue} {0.95}    & \textcolor{red} {\textbf{1.29}} \bigstrut[t]\\
    $\Delta$SSIM   &  0.009     &   0.009    & 0.012      &    0.011   &   0.013    &   0.016    &   \textcolor{blue} {0.017}    &  \textcolor{red} {\textbf{0.021}} \bigstrut[b]\\ 
    \hline
    \end{tabular}%
  \label{tab:table2}%
  \vspace{-8pt}
\end{table*}%

\textbf{Datasets.}
We train and test the proposed video compression framework respectively on LDV 2.0 \cite{yang2022ntire} and MFQE 2.0 \cite{guan2019mfqe}. LDV 2.0 consists of 270 videos, 255 for the training set, and 15 for the test set. They are all compressed using the official HEVC test model (HM 16.20) at QP = 37 the default Low-Delay P (LDP) setting. MFQE 2.0 consists of 126 videos, 108 for the training set, and 18 for the test set. In MFQE 2.0, the videos are compressed by HM 16.5 at LDP mode with QP = 37.

\vspace{3pt}
\noindent
\textbf{Implementation Details. }
We adopt Adam optimizer \cite{kingma2014adam} and Cosine Annealing scheme \cite{loshchilov2016sgdr}. We use L2 loss as the intermediate loss function and Charbonnier loss as the main loss function. The initial learning rate of the main network and the ﬂow network are set to $1\times10^{-4}$ and $2.5\times10^{-5}$, respectively. The total number of iterations is 300K. The weights of the ﬂow network are ﬁxed during the ﬁrst 5,000 iterations, and the weights of the LDR-AE encoder network are ﬁxed during the ﬁrst 20,000 iterations. The overall training comprises 300,000 iterations, during which training with intermediate supervision is conducted for 150,000 iterations. The batch size is 2 and the patch size of input compressed frames is $256\times256$. The number of denoising blocks is set to 4. The number of feature channels is 64. Detailed experimental settings and model architectures are provided in the supplementary material.

\subsection{Comparisons with State-of-the-Art Methods}

\begin{figure*}
  \centering
  \includegraphics[width=1.00\textwidth]{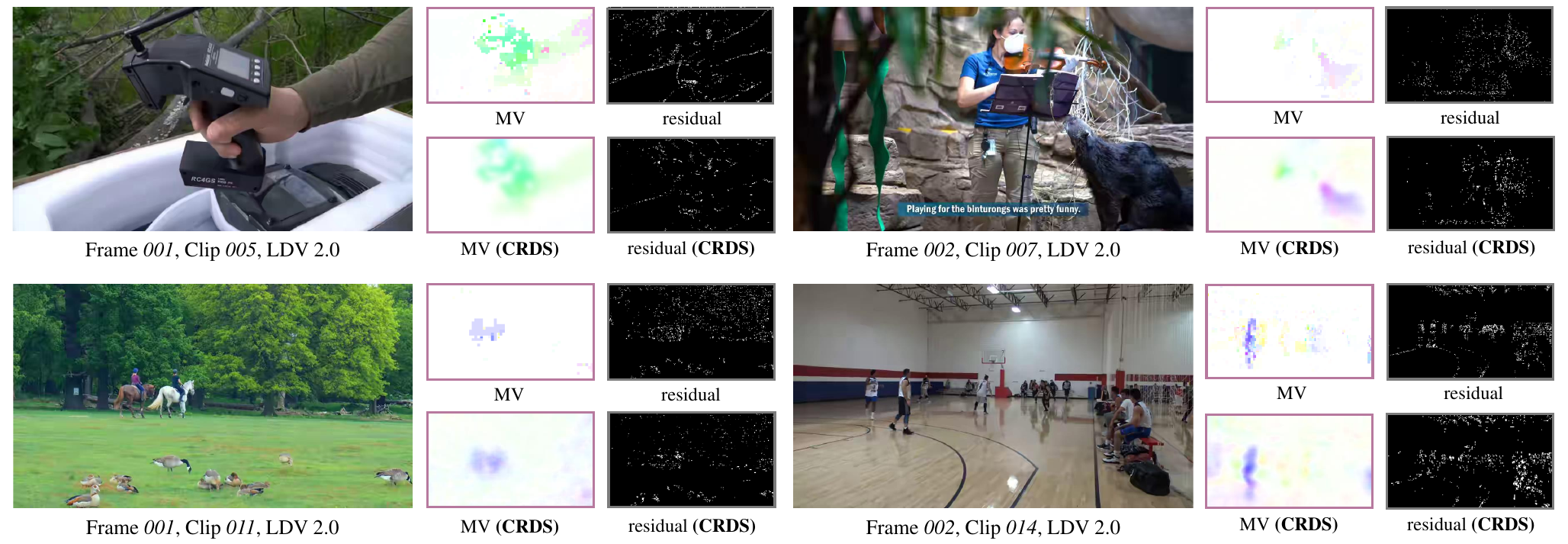} 
  \caption{Visualization of motion vectors and residuals from x264 \cite{x264} codec and CRDS, where obvious similarity can be observed. }
  \label{fig:fig5}
\end{figure*}

We compare CRDS with 7 video restoration or enhancement models, as listed in Table \ref{tab:table1}, where $\#1\sim\#15$ refer to the 15 videos
comprising the LDV 2.0 test set. We also provide the parameters and runtime of different models.
\begin{itemize}[leftmargin=0pt, itemindent=2ex, topsep=1pt, itemsep=0pt, partopsep=0pt]
\item
\textbf{IID scenario.}
As shown in Table \ref{tab:table1}, CRDS reaches the forefront in performance on the LDV 2.0 and MFQE 2.0 datasets. The previous state-of-the-art networks available for the VQE task have only managed to achieve a maximal increase of 1.73 in PSNR. In contrast, our CRDS network is the first to achieve $\Delta PSNR > 2$ on the LDV 2.0, surpasses the previously leading BasicVSR++ by as much as 0.33 dB in PSNR on LDV 2.0 and 0.23 dB on MFQE 2.0, with the parameters and runtime slightly exceed those of BasicVSR++.

As depicted in Figure \ref{fig:fig4}, video compression can cause a visibly significant quality decline in challenging scenarios such as low-light environments, during rapid motion, and in scenes with complex textures. The qualitative comparisons shown in Fig. \ref{fig:fig4} demonstrate that CRDS effectively restores fine details. Even in scenarios where fast-moving objects result in blurring within a video frame, CRDS significantly enhances the quality of the compressed image. For Frame \textit{002}, Clip \textit{015} on LDV 2.0, the video compression caused blurring in the area where the hand intersects with the ground. EDVR and VRT struggle to recover the texture of the ground, while the video enhancements by BasicVSR and IconVSR make it difficult to distinguish between the hand and the ground. CRDS, on the other hand, best restores the texture of the ground and also maintains the distinction between the hand and the ground.

\begin{table}[tbp]
\footnotesize
  \centering
  \caption{Ablation study of MuNA residual extraction, pre-trained LDR auto-encoder and intermediate supervision on LDV 2.0 \cite{yang2022ntire}. [Key: \textbf{Best}].}
  \vspace{-8pt}
    \begin{tabular}{l|p{0.9cm}<{\centering}|p{0.9cm}<{\centering}|p{0.9cm}<{\centering}|p{0.9cm}<{\centering}|p{1.0cm}
    <{\centering}}
    \toprule
          & \multicolumn{1}{c|}{(A)} & \multicolumn{1}{c|}{(B)} & \multicolumn{1}{c|}{(C)} & \multicolumn{1}{c|}{(D)} & \multicolumn{1}{c}{\textbf{CRDS}} \\
    \midrule
    MuNA &       &   \checkmark    &   \checkmark    & \checkmark    & \checkmark  \\
    LDR Encoder &       &       &   \checkmark    & \checkmark    & \checkmark  \\
    Interm. Supv. &       &       &       &   \checkmark    & \checkmark \\
    LDR Decoder  &       &       &       &     & \checkmark \\
    \midrule
    $\Delta$PSNR (dB) &  1.69     &  1.92     &   1.97    &  2.04  &  \textbf{2.06}\\ 
    \midrule
    $\Delta$SSIM &   0.036    &   0.046    &  0.047   &  0.050   & \textbf{0.051} \\
    \bottomrule
    \end{tabular}%
  \label{tab:table3}%
  \vspace{-7pt}
\end{table}%

\begin{table}[tbp]
\vspace{-2pt}
\footnotesize
  \centering
  \caption{Performance comparison between different motion estimation and feature alignment methods. [Key: \textbf{Best}].}
  \vspace{-8pt}
    \begin{tabular}{l|p{1.5cm}<{\centering}|p{1.5cm}<{\centering}|p{1.1cm}<{\centering}||p{1.1cm}
    <{\centering}}
    \toprule Method
          & \multicolumn{1}{c|}{\tabincell{c}{Optical Flow\\ \cite{ranjan2017optical}}} & \multicolumn{1}{c|}{\tabincell{c}{Flow-Guided\\DCN \cite{chan2022basicvsr++}}} & \multicolumn{1}{c||}{\tabincell{c}{TMSA\\ \cite{liang2022vrt}}} & \multicolumn{1}{c}{\textbf{MuNA}}
          \\
    \midrule
    $\Delta$PSNR (dB) &   1.15    &   1.57    &   1.61    &    \textbf{1.92}      \\
    $\Delta$SSIM  &  0.027     &   0.033    &    0.035   &   \textbf{0.046 } 
       \\
    \bottomrule
    \end{tabular}%
  \label{tab:table4}%
  \vspace{-13pt}
\end{table}%

\item
\textbf{OOD scenario.}
\vspace{-2pt}
When trained on the MFQE 2.0 dataset and tested on the LDV 2.0 dataset, models with better generalizability are expected to perform more effectively under this Out-Of-Distribution (OOD) scenario, as the two datasets are not identically distributed. As shown in Table \ref{tab:table2}, the CRDS model significantly outperforms other models in this OOD scenario for video quality enhancement,  which can be attributed to the codec's inductive bias introduced by the network's structured design.

\end{itemize}

\subsection{Intermediate Results in Frame Prediction}
To demonstrate how the network architecture of CRDS aligns with a codec, we compared the intermediate outputs of corresponding modules. Visual results are shown in Figure \ref{fig:fig5}.
We first compared the motion vector output of MuNA-ME with the actual codec's motion vectors. It's observable that the outputs of MuNA-ME closely approximate the motion vectors generated by the codec. Additionally, we visualize the latent representation of residuals output by MuNA-AGP through channel summation and binarization, producing residuals very similar to those of the codec. These results validate the interpretability of the CRDS modules.

\subsection{Ablation Study}

To assess the impact of the proposed components, we initially establish a baseline and then incrementally integrate each component. As evident from Table \ref{tab:table3}, every individual component contributes significantly to overall performance enhancement.
\begin{itemize}[leftmargin=0pt, itemindent=2ex, topsep=1pt, itemsep=0pt, partopsep=0pt]
\item
\textbf{Mutual Neighborhood Attention. }
As shown in Table \ref{tab:table3}, incorporating MuNA for residual extraction in the denoising block results in an improvement of 0.23 dB. We further compare different methods for motion estimation and feature alignment in Table \ref{tab:table4}, showing that our proposed MuNA is the most effective in utilizing spatial-temporal information of the current and adjacent frames.

\item
\textbf{Pre-trained LDR-AE. }
It is demonstrated in Table \ref{tab:table3}  that utilizing a pre-trained LDR-AE as the encoder for our network and freezing it as initial weights during the initial training phase can yield a 0.05 dB performance enhancement. Furthermore, employing the pre-trained LDR-AE to provide initial weights for the lightweight decoder in PDIS intermediate supervision also results in an additional 0.02 dB improvement compared to random initialization.

\item
\textbf{Intermediate supervision via progressive hybrid targets. }
For our multi-block network architecture with progressive denoising, employing our proposed method of progressively mixing original and compressed frames to provide intermediate targets during the initial stages of network training will result in a 0.07 dB improvement if the lightweight decoder for intermediate supervision is initialized randomly, as shown in Table \ref{tab:table3}.

\end{itemize}

\section{Conclusion}
We have presented a structural deep model for video quality enhancement, which introduces compression-related inductive biases into black-box neural networks by aligning the network structure with the key components of the video compression process. The experiments on two video quality enhancement benchmarks show that CRDS achieves state-of-the-art performance. We believe the design methodology of the proposed deep structural model can also extend to other video-related low-level tasks that require prior knowledge.

{\small
\bibliographystyle{unsrt}
\bibliography{sample-base}
}

\end{document}